\documentclass{article}
\usepackage{amsmath,amsthm,amssymb,amsfonts,mathtools,bm,bbm,array,float,mathdots,dsfont,mathrsfs}
\usepackage{caption,subcaption,multirow,longtable,lscape,wrapfig,tikz,comment}
\usepackage{booktabs}
\usepackage{hyperref}
\hypersetup{allcolors=black}
\usepackage{graphicx,tikz,tikz-cd}
\usetikzlibrary{positioning}
\tikzset{
  node/.style={circle, draw=blue!70!black, fill=blue!20, thick, minimum size=4.5mm, inner sep=0pt},
  edge/.style={->, line width=0.9pt, >=Latex, shorten >=1pt, shorten <=1pt},
  every label/.style={font=\small}
}
\usepackage[top=1in, bottom=1in, left=1in, right=1in]{geometry}
\usepackage[absolute]{textpos}
\usepackage[toc,page]{appendix}
\usepackage[utf8]{inputenc}
\usepackage[autostyle=true]{csquotes}
\allowdisplaybreaks

\def\bea{\begin{eqnarray}}
\def\eea{\end{eqnarray}}

\newcommand{\ghlogo}{\raisebox{-0.2em}{\includegraphics[height=1em]{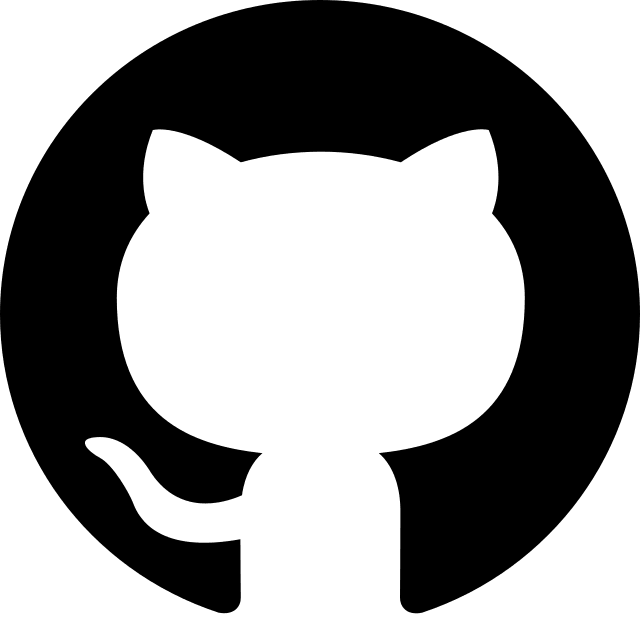}}}

\newcommand{\tabSetup}{%
\begin{tabular}{llrrr}
\toprule
Architecture & Dataset & Parameters & Epochs & Dense accuracy [\%] \\
\midrule
SimpleNN & MNIST & 55\,050 & 50 & $97.59 \pm 0.27$ \\
SimpleNN & CIFAR-10 & 201\,482 & 50 & $50.46 \pm 0.63$ \\
SimpleViT & MNIST & 539\,914 & 50 & $98.70 \pm 0.10$ \\
SimpleViT & CIFAR-10 & 545\,930 & 50 & $81.22 \pm 0.47$ \\
\bottomrule
\end{tabular}%
}

\newcommand{\tabMain}{%
\begin{tabular}{lcccc c}
\toprule
Scheme & AUC (acc.) & floor-corr. & AUC (MCC) & vs.\ mag.\ (acc.\,/\,MCC) & Compute \\
\midrule
\multicolumn{6}{l}{\textit{SimpleNN --- MNIST}} \\
\midrule
Magnitude & $0.907 \pm 0.007$ & $0.896 \pm 0.008$ & $0.898 \pm 0.007$ & $1.00\times$\,/\,$1.00\times$ & 6.3 s \\
FIM & $0.743 \pm 0.022$ & $0.714 \pm 0.024$ & $0.718 \pm 0.023$ & $0.82\times$\,/\,$0.80\times$ & 9.9 s \\
F-dist (one-shot) & $0.891 \pm 0.008$ & $0.879 \pm 0.009$ & $0.881 \pm 0.009$ & $0.98\times$\,/\,$0.98\times$ & 9.9 s \\
F-dist (iterative) & $0.924 \pm 0.003$ & $0.916 \pm 0.003$ & $0.916 \pm 0.003$ & $1.02\times$\,/\,$1.02\times$ & 9.9 s \\
F-dist (global) & $0.930 \pm 0.002$ & $0.922 \pm 0.002$ & $0.922 \pm 0.002$ & $1.02\times$\,/\,$1.03\times$ & 17.0 s \\
F-dist (exact) & $0.928 \pm 0.002$ & $0.920 \pm 0.002$ & $0.921 \pm 0.002$ & $1.02\times$\,/\,$1.03\times$ & 78.9 s \\
\midrule
\multicolumn{6}{l}{\textit{SimpleNN --- CIFAR-10}} \\
\midrule
Magnitude & $0.633 \pm 0.011$ & $0.542 \pm 0.013$ & $0.552 \pm 0.016$ & $1.00\times$\,/\,$1.00\times$ & 6.6 s \\
FIM & $0.634 \pm 0.008$ & $0.543 \pm 0.011$ & $0.559 \pm 0.006$ & $1.00\times$\,/\,$1.01\times$ & 10.0 s \\
F-dist (one-shot) & $0.890 \pm 0.005$ & $0.863 \pm 0.006$ & $0.867 \pm 0.006$ & $1.41\times$\,/\,$1.57\times$ & 10.1 s \\
F-dist (iterative) & $0.893 \pm 0.004$ & $0.866 \pm 0.004$ & $0.869 \pm 0.004$ & $1.41\times$\,/\,$1.58\times$ & 10.0 s \\
F-dist (global) & $0.892 \pm 0.005$ & $0.865 \pm 0.005$ & $0.869 \pm 0.005$ & $1.41\times$\,/\,$1.57\times$ & 17.0 s \\
F-dist (exact) & $0.892 \pm 0.004$ & $0.866 \pm 0.005$ & $0.869 \pm 0.005$ & $1.41\times$\,/\,$1.58\times$ & 6.3 min \\
\midrule
\multicolumn{6}{l}{\textit{SimpleViT --- MNIST}} \\
\midrule
Magnitude & $0.764 \pm 0.005$ & $0.738 \pm 0.005$ & $0.743 \pm 0.006$ & $1.00\times$\,/\,$1.00\times$ & 5.9 s \\
FIM & $0.565 \pm 0.030$ & $0.517 \pm 0.033$ & $0.525 \pm 0.031$ & $0.74\times$\,/\,$0.71\times$ & 35.8 s \\
F-dist (one-shot) & $0.783 \pm 0.017$ & $0.759 \pm 0.019$ & $0.765 \pm 0.021$ & $1.03\times$\,/\,$1.03\times$ & 35.7 s \\
F-dist (iterative) & $0.865 \pm 0.008$ & $0.850 \pm 0.008$ & $0.851 \pm 0.009$ & $1.13\times$\,/\,$1.14\times$ & 36.0 s \\
F-dist (global) & $0.837 \pm 0.008$ & $0.819 \pm 0.009$ & $0.822 \pm 0.008$ & $1.10\times$\,/\,$1.11\times$ & 1.6 min \\
F-dist (exact) & $0.867 \pm 0.008$ & $0.853 \pm 0.009$ & $0.854 \pm 0.010$ & $1.14\times$\,/\,$1.15\times$ & 3.38 d$^\dagger$ \\
\midrule
\multicolumn{6}{l}{\textit{SimpleViT --- CIFAR-10}} \\
\midrule
Magnitude & $0.678 \pm 0.008$ & $0.633 \pm 0.009$ & $0.636 \pm 0.009$ & $1.00\times$\,/\,$1.00\times$ & 6.9 s \\
FIM & $0.484 \pm 0.014$ & $0.412 \pm 0.016$ & $0.440 \pm 0.017$ & $0.71\times$\,/\,$0.69\times$ & 37.5 s \\
F-dist (one-shot) & $0.751 \pm 0.015$ & $0.716 \pm 0.016$ & $0.729 \pm 0.014$ & $1.11\times$\,/\,$1.15\times$ & 37.4 s \\
F-dist (iterative) & $0.791 \pm 0.013$ & $0.762 \pm 0.014$ & $0.770 \pm 0.014$ & $1.17\times$\,/\,$1.21\times$ & 37.0 s \\
F-dist (global) & $0.759 \pm 0.013$ & $0.725 \pm 0.015$ & $0.736 \pm 0.016$ & $1.12\times$\,/\,$1.16\times$ & 1.6 min \\
F-dist (exact) & $0.791 \pm 0.014$ & $0.761 \pm 0.015$ & $0.770 \pm 0.015$ & $1.17\times$\,/\,$1.21\times$ & 4.59 d \\
\bottomrule
\end{tabular}
}

\newcommand{\tabK}{%
\begin{tabular}{ccccc}
\toprule
$K$ & probes/coord. & AUC (acc.) & $\Delta$ vs.\ $K\!=\!2$ & Compute \\
\midrule
2 & 1 & $0.9283 \pm 0.0020$ & $+0.0000$ & 57.3 s \\
3 & 2 & $0.9285 \pm 0.0015$ & $+0.0002$ & 78.9 s \\
5 & 4 & $0.9272 \pm 0.0033$ & $-0.0010$ & 1.9 min \\
9 & 8 & $0.9292 \pm 0.0032$ & $+0.0010$ & 3.5 min \\
17 & 16 & $0.9269 \pm 0.0042$ & $-0.0014$ & 5.5 min \\
\bottomrule
\end{tabular}%
}

\begin{document}
\begin{textblock}{5}(12,1)
\noindent QMUL-PH-26-32
\end{textblock}
\title{\bf Optimal Pruning for Neural Architectures \\using Fisher Information Distances \\ }
\author{\bf David S. Berman\textsuperscript{1}\footnote{d.s.berman@qmul.ac.uk}, Yen-Yu Fu\textsuperscript{1}\footnote{jorden15937@gmail.com}, Edward Hirst\textsuperscript{2}\footnote{ehirst@unicamp.br}, Thelma Chiwete Obirai\textsuperscript{1}\footnote{thelmaobirai@gmail.com\\ Authors listed alphabetically.}}
\date{\small\today\\ \mbox{}\\ \textsuperscript{1} \textit{\small Centre for Theoretical Physics, \\Queen Mary University of London,\\ Mile End Road, London E1 4NS, UK}\\
\mbox{}\\
\textsuperscript{2} \textit{\small Institute of Mathematics, Statistics and Scientific Computing,\\ University of Campinas (Unicamp),\\ 13083-859, Brazil}\\
}

\maketitle

\begin{abstract}
A new scheme for parameter pruning is introduced, derived from the differential-geometric distance in model space.
Pruning a parameter sets its value to zero, representing a displacement of the model to the hypersurface on which that parameter vanishes.
The minimal distance from the unpruned model to this hypersurface is naturally computed via the geodesic distance in the model space as determined by the Fisher information metric. This distance determines the true change in the model, and its performance, under pruning.
By analysing progressively more faithful approximations of this geodesic distance a natural hierarchy of optimality for pruning methods is determined. This starts with the traditional magnitude pruning, then develops into new more sophisticated and effective pruning schemes.
The method is demonstrated for both fully-connected networks and vision transformers, on MNIST and CIFAR-10, over the complete $0$--$100\%$ pruning range and across five random seeds.
It outperforms pruning by parameter magnitude and by the local Fisher information alone in every architecture and dataset combination considered, on both accuracy and the Matthews correlation coefficient.
Additionally, analysis of different levels of geodesic approximation produces intermediate pruning schemes that are computationally efficient and maintain near-optimal performance.

This geometric picture supplies not only a state-of-the-art pruning methodology for AI models, but also a verified and mathematically-motivated justification for pruning schemes.

\mbox{} \\ 
\small \ghlogo: \, \href{https://github.com/edhirst/Fdist_Pruning}{\texttt{https://github.com/edhirst/Fdist\_Pruning}}
\end{abstract}

\thispagestyle{empty}
\clearpage
\setcounter{page}{1}
\numberwithin{equation}{section}
\numberwithin{figure}{section}
\numberwithin{table}{section}
\tableofcontents
\clearpage 

\section{Introduction}\label{sec:intro}

Trained neural architectures are overwhelmingly overparameterised.
A large fraction of their weights may be removed without measurable loss of performance, and the practical value of doing so (in memory, latency, energy) has made \emph{pruning} standard in the deployment pipeline \cite{Han:2015pru, Han:2016dc, Blalock:2020state}.
Every pruning method rests on an \emph{importance measure}: a scalar assigned to each parameter, ranking the order in which parameters are discarded.
Since the quality of the pruned model is decided almost entirely by that ranking, the choice of measure is central to ensuring effective pruning.

To date, two methodologies dominate industry practice.
The first, and by far the most widely used, is the parameter magnitude $|\theta^k|$ \cite{Han:2015pru, Frankle:2019ltt}. A weight close to zero is presumed not to matter when acting under the assumption that setting it to zero changes the model only slightly.
The second appeals to the local curvature of the log-likelihood. In a lineage stretching from the "Optimal Brain Damage" and "Optimal Brain Surgeon" works \cite{NIPS1989_6c9882bb, Hassibi:1992obs} through to the Fisher-based criteria of \cite{Theis:2018fisher, Molchanov:2017prune}, a weight to which the model output is insensitive is presumed to matter little, whatever its size.
This work demonstrates that both approaches are only partially correct  within a larger, more accurate and effective framework.
The first measures a displacement, but in the Euclidean metric on parameter space (a metric with no statistical meaning, and one which is not even invariant under reparameterisation of the model).
The second measures sensitivity only \emph{at} the trained point, and says nothing about how much the parameter must change in order to zero it and hence remove it.

The resolution is that pruning is not an infinitesimal perturbation at all.
To prune the parameter $\theta^k$ is to move the model a \emph{finite} distance, from its trained value $(\theta^\ast)^k$ to the hypersurface $\theta^k = 0$.
What one wishes to make small is the change this induces in the \emph{distribution} the network represents, a question which information geometry \cite{Rao:1945, Cencov:1982, Amari:2000ig} provides a natural means for assessing through the Fisher information matrix (FIM).
The Fisher information matrix is the canonical Riemannian metric on the space of models \cite{Amari:1998ng}, and the distance between two models is the length of the shortest path joining them, obtained by integrating that metric along that path.
This work's central hypothesis is that the importance of a parameter is determined by its \emph{Fisher geodesic distance} to zero. This comes from a natural differential-geometric interpretation of the model space. 

The derivation of \S\ref{sec:fdist_prune} is instructive in its own right.
Under the gross simplifying assumptions that the Fisher metric is diagonal, that the relevant path is the coordinate line along $\theta^k$, and that the metric is constant along it, the Fisher distance collapses to,
\begin{equation}
    d_I = \sqrt{I_{kk}}\,\big|(\theta^\ast)^k\big|,
\end{equation}
the product of the magnitude criterion and the (square-rooted) Fisher criterion.
The two standard measures are thus recovered as factors of a single geometric quantity once the simplifying assumptions have been made. 
The presentation as competing heuristics is simply an artefact of ignoring one of the factors while assuming the requisite simplifying assumptions. 
There are thus two immediate improvements. 
The first is to consider the full Fisher distance and second is to relax the constancy assumption. 
One may then generate a hierarchy of successively better approximations to the true Fisher distance by recomputing the metric at each pruning step and then resolving it at $K$ points along the path itself.

This hierarchy is evaluated against the magnitude and Fisher baselines for two architectures (a fully-connected network and a vision transformer), on two standard datasets (MNIST and CIFAR-10), sweeping the entire $0$--$100\%$ pruning range and averaging over five random seeds.
Three findings emerge.
First, the Fisher-distance family of pruning methods consistently outperforms both magnitude and local-FIM baseline methods in all four architecture--dataset combinations, on accuracy and on the Matthews correlation coefficient alike.
Second, the Fisher criterion by itself is \emph{worse} than the magnitude criterion in three of the four, and markedly so on the transformer, which shows the advantage cannot be attributed to second-order information, but to its combination with displacement.
Third, and least expected, the most accurate approximation of the geodesic distance often provides negligible improvement, such that within this new family of approximations the more easily computed (cheaper) approximants perform equally well as the best approximation. 

To emphasise, the computational efficiency of the low-level approximations means there is real potential for efficient pruning at scale. 
This geometric picture consequently delivers two things: a better importance measure for model parameters and a series of efficient and effective cheaper approximations.

\newpage
The paper is organised as follows.
Section \S\ref{sec:background} collects the background on the architectures considered, Fisher information and the geometry it induces on model space, and the existing standard pruning schemes against which these new methods are compared. Section \S\ref{sec:fdist_prune} derives the Fisher-distance measure and the hierarchy of approximations to it. Section \S\ref{sec:numerics} presents the numerical results and analysis, and section \S\ref{sec:summary} summarises these.
Work is completed in \texttt{python}, using \texttt{pytorch} \cite{Paszke:2019pytorch}, and scripts are made available at this work's respective \href{https://github.com/edhirst/Fdist_Pruning}{\texttt{GitHub}}\footnote{\href{https://github.com/edhirst/Fdist_Pruning}{\texttt{https://github.com/edhirst/Fdist\_Pruning}}} repository.

\section{Background}\label{sec:background}

\subsection{Neural Architectures}\label{sec:nns}

A neural network for classification is a parametric map $f_\theta : \mathcal{X} \to \mathbb{R}^C$ from inputs to $C$ class logits, whose parameters $\theta \in \mathbb{R}^P$ are fit by minimising a loss over a training set.
Composing $f_\theta$ with a softmax makes the network a conditional probability model,
\begin{equation}
\label{eq:softmax_model}
p(y = c \mid x, \theta)
=
\frac{\exp\big(f_\theta(x)_c\big)}{\sum_{c'=1}^{C}\exp\big(f_\theta(x)_{c'}\big)}\,,
\end{equation}
and it is this interpretation of the network (as a point in a space of probability distributions, rather than as a function) that the geometry of \S\ref{sec:infogeo} requires.
The parameter vector $\theta$ is a coordinate system on that space, and pruning is finite motion within it.

Two architectures are considered, chosen to span the gap between the simplest useful model and a contemporary attention-based one while remaining small enough that the exact computation of \S\ref{sec:fdist_prune} was affordable.

The first, denoted \texttt{SimpleNN}, is a fully-connected network: the input image is flattened and passed through two hidden layers of width $64$ with \texttt{ReLU} activations to a $C=10$ output.
Its parameter count is set by the input dimension, giving $55{,}050$ parameters on $28\times28$ greyscale data (MNIST) and $201{,}482$ on $32\times32$ colour data (CIFAR).

The second, denoted \texttt{SimpleViT}, is a compact vision transformer \cite{Dosovitskiy:2021vit} of the type designed for training from scratch on small datasets \cite{Hassani:2021cct}.
Images are divided into $4\times4$ patches by a strided convolution, embedded in dimension $128$, and augmented with a learned positional embedding; four pre-norm transformer blocks \cite{Vaswani:2017attn} with four attention heads each follow, and the token representations are mean-pooled before a linear classification head.
This gives approximately $5.4\times10^{5}$ parameters.
Following standard practice in transformer sparsity, the \texttt{LayerNorm} parameters and the positional embedding (together some $1.9\%$ of the total) are excluded from pruning, though reported pruning ratios remain fractions of \emph{all} parameters so that the sweeps are directly comparable between architectures.

\subsection{Datasets}\label{sec:data}

The two datasets used are MNIST \cite{LeCun:1998mnist}, $28\times28$ greyscale handwritten digits, and CIFAR-10 \cite{Krizhevsky:2009cifar}, $32\times32$ colour natural images; both are ten-class balanced classification problems with $60{,}000$ training and $10{,}000$ test examples.
Each is used with each architecture, giving four testbeds.

The pairing is deliberate rather than merely conventional.
MNIST is easy enough that a small network retains essentially all of its accuracy until a large majority of its weights have been removed, so that competing importance measures are indistinguishable across most of the pruning range and separate only in the extreme tail.
CIFAR-10 degrades gradually under pruning for both architectures, and it is there that the schemes can be compared over the whole of the $0$--$100\%$ range.
Including both highlights this ceiling effect, as reported in \S\ref{sec:results_main}.

\subsection{Fisher Information}\label{sec:fisher_info}

Let $p(x\,|\,\theta)$ be a parametric family of probability densities where $\theta\in\mathbb{R}^n$ is a vector of model parameters for this parametric family.
Then, the Fisher Information Matrix (FIM) $I(\theta)\in\mathbb{R}^{n\times n}$ quantifies the curvature of the log-likelihood and induces a natural Riemannian metric on the parameter manifold for the model space spanned by $\theta$.
The FIM may be defined in two ways:

\begin{equation}
\label{eq:FIM_form1}
I_{ij}(\theta)
=
\mathbb{E}_{(x,y)\sim\mathcal{D}}
\Bigg[
\sum_{b=1}^{B}\sum_{c=1}^{C}
\bigg(\frac{\partial}{\partial\theta^{i}}\log p\big(y_{b,c}\mid x_b,\theta\big)\bigg)
\bigg(\frac{\partial}{\partial\theta^{j}}\log p\big(y_{b,c}\mid x_b,\theta\big)\bigg)
\Bigg],
\end{equation}

\begin{equation}
\label{eq:FIM_form2}
I_{ij}(\theta)
=
-\,\mathbb{E}_{(x,y)\sim\mathcal{D}}
\Bigg[
\sum_{b=1}^{B}\sum_{c=1}^{C}
\frac{\partial^2}{\partial\theta^{i}\partial\theta^{j}}
\log p\big(y_{b,c}\mid x_b,\theta\big)
\Bigg].
\end{equation}
For an (input, output) pair $(x,y)$ drawn from a data distribution $\mathcal{D}$, where $b$ runs over the finite input data batch (which would be an integral over the full data space), and $c$ the output coordinates (i.e. the class indices for classification problems, or the output dimensions for regression problems).

The FIM may be written in either of these forms as: 
\begin{enumerate}
    \item The expected outer-product of scores, \eqref{eq:FIM_form1}.
    \item The negative expected Hessian of the log-likelihood, \eqref{eq:FIM_form2}.
\end{enumerate}
where these two forms are identical under mild regularity conditions such that integration and differentiation
may be interchanged (via integration by parts methods with a suitable exact term) and the score has zero mean.

There is a further distinction, immaterial to the definition but essential in practice, in how the expectation over outputs is taken.
Replacing $y$ by the labels actually observed in the data gives the \emph{empirical} Fisher; taking the expectation over the model's own predictive distribution, $c \sim p(\cdot \mid x, \theta)$, gives the \emph{model} Fisher,
\begin{equation}
\label{eq:model_fisher}
I_{ij}(\theta)
=
\mathbb{E}_{x\sim\mathcal{D}}\,
\mathbb{E}_{c\sim p(\cdot\mid x,\theta)}
\Bigg[
\frac{\partial \log p(c\mid x,\theta)}{\partial\theta^{i}}
\frac{\partial \log p(c\mid x,\theta)}{\partial\theta^{j}}
\Bigg]\,.
\end{equation}
Only the latter is the Fisher information of the model, and only the latter carries the geometric meaning that \S\ref{sec:infogeo} relies on; the empirical Fisher coincides with it at a perfectly fitted optimum but not in general, and is known to behave poorly as a curvature proxy away from one \cite{Kunstner:2019ef}.
Therefore \eqref{eq:model_fisher} is used throughout.

The matrix $I_{ij}$ has $P^2$ entries, which for even the modest architectures of \S\ref{sec:nns} is prohibitive to form, let alone to invert or to evaluate repeatedly along a path.
Here, the standard practice of the pruning and continual-learning literatures \cite{NIPS1989_6c9882bb, Kirkpatrick:2017ewc, Theis:2018fisher} in retaining only the diagonal is followed,
\begin{equation}
\label{eq:diag_approx}
I_{ij}(\theta) \;\approx\; I_{ii}(\theta)\,\delta_{ij}\,,
\end{equation}
which reduces the cost to $\mathcal{O}(P)$ and, in the geometric language of the next subsection, amounts to treating the coordinate directions of parameter space as mutually orthogonal.
This is an approximation and not a mild one; structured alternatives which retain block correlations, such as the Kronecker-factored approximation \cite{Martens:2015kfac}, exist and would refine every measure considered here; this point is returned to in \S\ref{sec:summary}.

\subsection{Information Geometry of Model Space}\label{sec:infogeo}

The reason to prefer the Fisher information over any other quadratic form on parameter space is that it is, in a precise sense, the only natural choice.

The first indication of this is that one does not have to posit it at all, it is what the canonical measure of statistical separation produces on its own.
Pruning exchanges one distribution in the family $\{p(\cdot\mid\theta)\}$ for another, and the standard measure of how far apart two such distributions lie is the Kullback--Leibler divergence \cite{Kullback:1951},
\begin{equation}
\label{eq:kl}
D_{\rm KL}\big(p_{\theta}\,\|\,p_{\theta'}\big)
=
\mathbb{E}_{x\sim p(\cdot\,|\,\theta)}
\bigg[\log\frac{p(x\,|\,\theta)}{p(x\,|\,\theta')}\bigg]\,,
\end{equation}
which is not itself a distance, being neither symmetric in its arguments nor subject to the triangle inequality.
It does, however, determine one.
Setting $\theta' = \theta + \delta\theta$ and expanding, the constant term vanishes because $D_{\rm KL}(p_\theta\,\|\,p_\theta) = 0$, and the linear term vanishes because the score has zero mean; equivalently, because a non-negative quantity which vanishes at $\theta' = \theta$ is stationary there.
The leading behaviour is therefore quadratic, and its coefficient is precisely \eqref{eq:FIM_form2},
\begin{equation}
\label{eq:kl_expansion}
D_{\rm KL}\big(p_{\theta}\,\|\,p_{\theta+\delta\theta}\big)
=
\tfrac{1}{2}\,I_{ij}(\theta)\,\delta\theta^{i}\delta\theta^{j}
+
\mathcal{O}\big(\delta\theta^{3}\big)\,.
\end{equation}
The asymmetry of \eqref{eq:kl} thus enters only at cubic order, leaving a symmetric quadratic form on the tangent space at $\theta$.
The Fisher information is in this sense not one candidate metric among many, but the infinitesimal limit of statistical distinguishability itself.

The second indication is a uniqueness statement.
Regarding a parametric family $\{p(\cdot\mid\theta)\}$ as a manifold whose points are distributions and whose coordinates are the parameters $\theta$, the Fisher information defines a Riemannian metric on it \cite{Rao:1945}.
{\v{C}}encov's theorem \cite{Cencov:1982} establishes that this metric is unique up to scale among those invariant under sufficient statistics, such that any measure of distinguishability between nearby distributions that does not depend on how the data are represented must be the Fisher metric.
The line element
\begin{equation}
\label{eq:line_element}
ds^2 = I_{ij}(\theta)\,d\theta^i\,d\theta^j
\end{equation}
therefore measures how distinguishable two infinitesimally separated models are, rather than how far apart their parameter values happen to be written.

Two consequences matter for what follows.
The first is invariance: $ds^2$ is unchanged by reparameterisation $\theta \to \tilde{\theta}(\theta)$, whereas the Euclidean length $\delta_{ij}d\theta^i d\theta^j$ implicit in the magnitude criterion is not.
A pruning importance built from \eqref{eq:line_element} is thus a statement about the model, while one built from parameter magnitudes is partly a statement about the arbitrary coordinates in which the model was written down.
The second is that a metric supplies distances between \emph{finite}, not merely infinitesimal, separations, by integration along a path; and it is exactly a finite separation, from $(\theta^\ast)^k$ to $0$, that pruning effects.
This is the observation that \S\ref{sec:fdist_prune} develops.

The same geometry underlies the natural gradient \cite{Amari:1998ng}, in which the steepest-descent direction is taken with respect to \eqref{eq:line_element} rather than the Euclidean metric, and it has been applied to the analysis of learning dynamics and representation in a number of recent works \cite{Berman:2022mak, Berman:2022uov, Berman:2023rqb, Berman:2024pax, Howard:2024kfd, Manning-Coe:2025uhs}.

\subsection{Existing Pruning Schemes}\label{sec:prune}

Pruning schemes may be considered to be classified by the importance measure $s_k$ they assign to each parameter $\theta^k$.
The scheme then removes parameters in increasing order of $s_k$, either in a single step or in iterative stages.
This section describes the two standard schemes for pruning known in the literature, and most often used in practical implementations to date. 
These methods are the baseline for comparison against the new class of Fisher-distance measures introduced in this work.

\subsubsection{Magnitude Pruning}\label{sec:mag_prune}

The magnitude criterion takes the importance of a parameter to be its absolute value,
\begin{equation}
\label{eq:mag_score}
s^{\rm mag}_k = \big|\theta^k\big|\,,
\end{equation}
and is the workhorse of the field \cite{Han:2015pru, Han:2016dc}.
Its appeal is that it is free (no gradients, no data) and that it works remarkably well, to the extent that surveys of the literature have found it difficult to demonstrate consistent improvement over it \cite{Blalock:2020state}.
Applied iteratively, with the network retrained between prunings, it is also the procedure that exposes the lottery-ticket phenomenon \cite{Frankle:2019ltt}.
It remains the basis of the state of the art at the largest scales, where retraining is unaffordable and one-shot criteria are all that can be applied \cite{Frantar:2023sparsegpt}.

The most successful modern refinement of \eqref{eq:mag_score} is instructive for what follows.
Wanda \cite{Sun:2024wanda} scores a weight not by its magnitude alone but by that magnitude multiplied by a data-dependent scale, the norm of the corresponding input activation,
\begin{equation}
\label{eq:wanda_score}
s^{\rm wanda}_{ij} = \big|\theta_{ij}\big| \cdot \big\|x_j\big\|_2\,,
\end{equation}
and this single change is enough to recover much of the performance of far more expensive second-order methods.
The structural lesson (that the useful quantity is a magnitude weighted by a measure of how much the model attends to that direction) is exactly the one the geometric derivation of \S\ref{sec:fdist_prune} arrives at from first principles, with the Fisher metric in place of the activation norm; as highlighted in \S\ref{sec:relation}.

The implicit justification for \eqref{eq:mag_score} is that setting a small parameter to zero is a small perturbation of the model.
This is a statement about distance, and as noted in \S\ref{sec:infogeo} it is a statement made in the Euclidean metric $\delta_{ij}$ on parameter space, treating a unit of displacement in every coordinate direction as equally consequential.
There is no reason for a trained highly non-linear network to satisfy this.
Directions in which the network is stiff and directions in which it is flat are weighted identically and a rescaling of any layer's parameters (which for many architectures leaves the function computed unchanged) reorders the ranking.
Magnitude pruning is thus best understood not as an unprincipled heuristic but as the correct geometric measure evaluated in the wrong metric.

\subsubsection{Fisher Pruning}\label{sec:fisher_prune}

The complementary family takes importance from the local curvature of the loss surface, and originates with Optimal Brain Damage \cite{NIPS1989_6c9882bb} and Optimal Brain Surgeon \cite{Hassibi:1992obs}.
Expanding the loss $L$ about a trained minimum, where the gradient vanishes, gives to second order
\begin{equation}
\label{eq:obd_expansion}
\delta L \;\simeq\; \tfrac{1}{2}\, H_{kk}\,\big(\delta\theta^k\big)^2
\end{equation}
for a perturbation of the single coordinate $\theta^k$, with $H$ the Hessian.
Since the Fisher information is the negative expected Hessian of the log-likelihood, \eqref{eq:FIM_form2}, it serves as a positive-semi-definite and cheaply computable stand-in for $H$, and the resulting criterion
\begin{equation}
\label{eq:fim_score}
s^{\rm fim}_k = I_{kk}
\end{equation}
underlies Fisher pruning as it is usually practised \cite{Theis:2018fisher, Molchanov:2017prune, Molchanov:2019importance}.
The same diagonal object is used as a parameter-importance weight in continual learning \cite{Kirkpatrick:2017ewc}, and closely related saliency scores drive pruning at initialisation \cite{Lee:2019snip, Wang:2020grasp, Tanaka:2020synflow}.

The criterion \eqref{eq:fim_score} corrects the deficiency of \eqref{eq:mag_score} (it is sensitive to the geometry) but introduces a complementary one.
It is a purely local quantity, evaluated at $\theta^\ast$, and it contains no reference to the distance the parameter must be moved.
A weight of large curvature but negligible magnitude is already almost pruned, and \eqref{eq:fim_score} will nonetheless protect it; a weight of modest curvature but large magnitude must travel far, and \eqref{eq:fim_score} is indifferent to that.
Furthermore, the expansion \eqref{eq:obd_expansion} is valid for infinitesimal $\delta\theta$, whereas the perturbation pruning actually applies, $\delta\theta^k = -(\theta^\ast)^k$, is often large.

The two criteria are therefore not rivals but fragments.
One measures displacement without the metric; the other measures the metric without the displacement.
The next section shows that the Fisher distance in model space contains both, as the leading term of a single expansion, and that carrying the expansion further yields measures which outperform either.

\subsection{Relation to This Work}\label{sec:relation}

Since the ingredients assembled here (the Fisher metric, the geometry of model space, and pruning) have each appeared in the literature, it is worth stating precisely what is and is not new in what follows.

The use of the Fisher information to organise model space, and specifically to identify which directions in parameter space a model can afford to lose, is not new; it is the basis of the Bayesian renormalization programme \cite{Berman:2022mak, Berman:2022uov, Berman:2023rqb, Berman:2024pax}, in which coarse-graining is performed with respect to an information-theoretic distinguishability scale set by the Fisher metric, and in which the eigenspectrum of the FIM supplies a measure of the relative importance of linear combinations of parameters.
That construction is global and spectral, it asks which directions in model space are resolvable at a given scale.
The measure developed here is local and metric, it asks, for one coordinate at a time, how far the model must travel to reach the hypersurface on which that coordinate vanishes.
The two are complementary readings of the same geometry, and the second is, to our knowledge, not one that has previously been made the explicit basis of a pruning criterion.

Nor is the product of a magnitude with a data-dependent weight new as a saliency score; \eqref{eq:wanda_score} is precisely of that form, as are the various Taylor-expansion criteria \cite{Molchanov:2019importance}.
What is new is the derivation, and geometric motivation, for this specific choice of weighting.
These scores are ordinarily proposed as heuristics and justified by their performance, whereas $\sqrt{I_{kk}}\,|\theta^k|$ arises here as the leading term of a well-defined geometric quantity, with the assumptions that produce it made explicit and each of them individually relaxable.
That is what makes the hierarchy of \S\ref{sec:estimators} possible: there is a defined object being approximated, so one can ask how good the approximation is, and (as \S\ref{sec:results_hierarchy} does) answer by computing the object exactly.

The Fisher--Rao distance between probability distributions is of course classical \cite{Rao:1945, Amari:2000ig}.
The contribution here is its application, in the restricted per-coordinate form that makes it computable, as a parameter importance measure for pruning, together with the empirical finding that the suitable cheaper approximations to it are still remarkably effective.

\section{Deriving the Fisher-Distance Pruning Scheme}\label{sec:fdist_prune}

The background of \S\ref{sec:background} supplies the two ingredients this section combines.
Pruning a parameter is a finite displacement in model space, from its trained value to zero, and model space carries a natural metric in which such a displacement has an unambiguous length.
An importance measure follows with no further assumption, in that the safest parameter to prune is the one whose removal moves the model the shortest distance.
\S\ref{sec:distance_measure} carries out that construction, reducing the Riemannian distance to a computable quantity through a sequence of approximations stated individually, so that each may afterwards be relaxed in turn; its leading term proves to be the product of the two standard criteria of \S\ref{sec:prune}.
\S\ref{sec:estimators} then collects those refinements into the four estimators compared in \S\ref{sec:numerics}, which differ in how faithfully they resolve the metric along the path and range in cost from $\mathcal{O}(1)$ to $\mathcal{O}(P)$ Fisher evaluations per pruning step.

\subsection{The Distance Measure}\label{sec:distance_measure}

Let $(\mathcal{M},g)$ be a smooth Riemannian manifold, parameterised by $p$, with metric tensor $g$.  
For a tangent vector $v\in T_{p}\mathcal{M}$, the quantity  $g(v,v)$ denotes the squared norm induced by $g$ at the point $p$.  
A smooth curve $\gamma\!:\,[0,1]\to\mathcal{M}$ connects $p$ and $p'$ for $\gamma(0)=p$ and $\gamma(1)=p'$, parameterised by $t$.  
Then the Riemannian distance $d_{g}(p,p')$ is defined as the infimum of the lengths of all such curves, where the length functional is given by the integral

\begin{equation}
\label{eq:riemann_distance}
d_{g}(p,p') = \inf_{\gamma:[0,1]\to\mathcal{M}} \left\{
\int_{0}^{1}
\sqrt{\,g_{ij}\big(\gamma(t)\big)\;
\frac{d\gamma^{i}(t)}{dt}\;
\frac{d\gamma^{j}(t)}{dt}\,}\; dt
\right\},
\end{equation}
where repeated indices \(i,j\) are summed (Einstein summation), \(g_{ij}\) is the metric tensor, and \(\gamma^{i}(t)\) are the coordinate components of the curve \(\gamma\).

Consider the manifold $\mathcal{M}$ to represent the model space of some AI architecture, parameterised by the model parameters $\theta$ such that the manifold becomes $\Theta$. 
Then distance in model space is defined using the FIM, $I_{ij}$ in place of the general Riemannian metric $g_{ij}$, such that

\begin{equation}
\label{eq:distance_FIM_metric}
d_{I}(\theta,\theta')
=
\inf_{\gamma:[0,1]\to\Theta}\left\{
\int_{0}^{1}
\sqrt{\,I_{ij}\big(\gamma\big)\;
\frac{d\gamma^{i}}{dt}\;
\frac{d\gamma^{j}}{dt}\,}\; dt
\right\}.
\end{equation}

Then let us use the assumption discussed in §\ref{sec:fisher_prune}, that the FIM has relatively negligible off-diagonal components, so may be assumed to be diagonal ($I_{ij} \sim 0$ for $i \neq j$).

\begin{equation}
\label{eq:distance_FIM_metric_diag}
\begin{split}
d_{I}(\theta,\theta')
& =
\inf_{\gamma:[0,1]\to\Theta}\left\{
\int_{0}^{1}
\sqrt{\,I_{ii}\big(\gamma\big)\;
\bigg(\frac{d\gamma^{i}}{dt}\bigg)^2}
\right\},\\
& = 
\inf_{\gamma:[0,1]\to\Theta}\left\{
\int_{0}^{1}
\sqrt{\,I_{ii}\big(\gamma\big)}\;
\bigg|\frac{d\gamma^{i}}{dt}\bigg|
\right\}.\\
\end{split}
\end{equation}
Now this computation is computationally infeasible for many reasons, the first being computing the integral for all possible curves to find the infimum.
Hence let us make a simplifying assumption, that by performing a pruning step and pruning one parameter, $\theta^k$, only that value is changed from the learnt value at $\theta^\ast$ which is $(\theta^\ast)^k$ to 0.
Then the assumption made is that along the infimum curve between points $\theta^\ast$ and $\theta'$, all coordinates are fixed except for one, $\theta^k$, such that $\frac{d\gamma^{i}(t)}{dt} = \delta^k_i \frac{d\theta^i(t)}{dt}$.
Therefore the distance measure simplifies to
\begin{equation}
\begin{split}
d_{I}(\theta,\theta')
& =
\int_{0}^{1}
\sqrt{\,I_{ii}\big(\theta(t)\big)}\;
\bigg|\frac{d\theta^{i}}{dt}\bigg| \delta^k_i dt\;,\\
& = 
\int_{0}^{1}
\sqrt{\,I_{kk}\big(\theta(t)\big)}\;
\bigg|\frac{d\theta^k}{dt}\bigg| dt\;,\\
\end{split}
\end{equation}
where the last line has applied the index summation with $\delta^k_i$ such that only the dimension $k$ terms are left.

Although more manageable, computing this distance still requires knowing the FIM at every point along this $\theta^k$ line.
The simplest assumption which can be made here is that the FIM is constant across the manifold, $I_{ij}(\theta) = I_{ij}$, which means it is constant throughout the integral and the distance reduces to 
\begin{equation}
\begin{split}
d_{I}\big((\theta^\ast)^k,\ \theta^k=0 \big)
& =
\int_{0}^{1}
\sqrt{\,I_{kk}}\;
\bigg|\frac{d\theta^k}{dt}\bigg| dt\;,\\
& = 
\sqrt{\,I_{kk}}\;
\int_{0}^{1}
\bigg|\frac{d\theta^k}{dt}\bigg| dt\;,\\
& = 
\sqrt{\,I_{kk}}\;
\int_{(\theta^\ast)^k}^{0} |d\theta^k|\;,\\
& = 
\sqrt{\,I_{kk}}\;
\int_{0}^{(\theta^\ast)^k} d\theta^k\;,\\
& = 
\sqrt{\,I_{kk}}\; (\theta^\ast)^k\;.
\end{split}
\end{equation}

This, quite conveniently, ends up being just the product of the measures from the Magnitude pruning ($|\Delta \theta| = (\theta^\ast)^k$) of §\ref{sec:mag_prune}, and the Fisher pruning (with the geometrically motivated square root) of §\ref{sec:fisher_prune}.

However, there are still opportunities to improve this approximation of the true distance in model space.
The first is the assumption that the FIM is constant throughout the model space manifold. This can be improved by just assuming it is constant along the line path of integration.
The formula is the same, but importantly now for the next prune step the FIM is recomputed for the model space end point of the last prune step.
Taking this further one could imagine partitioning the line path into $N$ segments, recomputing the FIM at the end of each segment, and summing these segment lengths.
If the segments were labelled $(\theta^k_0, \theta^k_1, ..., \theta^k_N)$, this would look like
\begin{equation}
\begin{split}
d_{I}(\theta^k_0,\ \theta^k_N) 
& = 
\sum_{n=0}^{N-1} d_{I}(\theta^k_n,\ \theta^k_{n+1}) \;,\\
& = 
\sum_{n=0}^{N-1} \sqrt{I_{kk}(\theta^k_n)}\ |\theta^k_{n+1}-\theta^k_n|\;.
\end{split}
\end{equation}

The estimator implemented in this work adopts exactly this partition, with a uniform grid.
Fixing a resolution $K \geq 2$, the coordinate line is sampled at the $K$ points $\alpha_j \theta^k$ with $\alpha_j = 1 - j/(K-1)$ for $j = 0,\ldots,K-1$, so that $\alpha_0 = 1$ is the current value of the weight and $\alpha_{K-1} = 0$ is the pruned one.
The $K-1$ segments so defined are of equal coordinate length $|\theta^k|/(K-1)$, so that the displacement is common to every segment and factors out of the sum; what remains is the magnitude of the weight multiplied by an average of the metric sampled along the path.
Averaging $I_{kk}$ over the $K$ nodes of the partition, and taking the square root once at the end, gives the measure evaluated in this work,
\begin{equation}
\label{eq:path_partition}
d_{I}\big(\theta^k,\ 0\big)
\;\approx\;
\sqrt{\frac{1}{K}\sum_{j=0}^{K-1}
I_{kk}\Big(\theta\big|_{\theta^k \to \alpha_j\theta^k}\Big)}
\;\;\big|\theta^k\big|\,,
\end{equation}
where the notation $\theta|_{\theta^k \to \alpha_j\theta^k}$ records that only the coordinate being scored is moved along the path, every other coordinate being held at its current value, as the single-coordinate assumption above requires.

Two properties of \eqref{eq:path_partition} should be stated plainly.
First, the square root is taken of the mean rather than the mean of the roots, which is not the literal discretisation of the length integral; the two coincide when $I_{kk}$ is constant along the path and differ at second order in its variation, and \S\ref{sec:results_hierarchy} reports the difference to be immaterial at the precision these experiments resolve.
Second, $K$ is a resolution parameter and not a modelling choice: raising it refines the quadrature of a fixed integral at a cost linear in $K$, so its influence is something the experiments can measure directly rather than something that must be assumed; \S\ref{sec:results_k} measures it.

\subsection{The Estimators Evaluated}\label{sec:estimators}

It is convenient to collect the approximations just derived into the explicit forms that are implemented and compared in \S\ref{sec:numerics}.
Writing $\theta$ for the current (partially pruned) parameter vector and $\theta^k$ for the coordinate whose importance is being scored, and taking the $K$ path points to be the uniform grid $\alpha \in \{1, \tfrac{K-2}{K-1}, \ldots, 0\}$ so that $\alpha\theta^k$ interpolates between the current value and zero as in \eqref{eq:path_partition}, the four members of the hierarchy are as follows.

\paragraph{One-shot (\texttt{f\_dist\_one\_shot}).}
The metric is evaluated once, at the trained point $\theta^\ast$, and held fixed for the whole pruning sweep:
\begin{equation}
\label{eq:score_oneshot}
s^{\rm one\text{-}shot}_k = \sqrt{I_{kk}(\theta^\ast)}\;\big|\theta^k\big|\,.
\end{equation}

\paragraph{Iterative (\texttt{f\_dist\_iterative}).}
The metric is recomputed on the current model at every pruning step, so that the local approximation is re-anchored as the model moves:
\begin{equation}
\label{eq:score_iterative}
s^{\rm iter}_k = \sqrt{I_{kk}(\theta)}\;\big|\theta^k\big|\,.
\end{equation}
This costs one Fisher evaluation per pruning step, and is otherwise identical to \eqref{eq:score_oneshot}.

\paragraph{Exact path (\texttt{f\_dist}).}
The metric is resolved along the coordinate line itself, each coordinate being perturbed on its own with all others held at their current values:
\begin{equation}
\label{eq:score_exact}
s^{\rm exact}_k
=
\sqrt{\frac{1}{K}\sum_{\alpha}\, I_{kk}\big(\theta\big|_{\theta^k \to \alpha\theta^k}\big)}\;\;\big|\theta^k\big|\,.
\end{equation}
This is \eqref{eq:path_partition}, evaluated on the current model at each pruning step, and is the quantity against which the cheaper members are judged.
Its cost is the reason it is rarely computed: a separate Fisher probe is required for every surviving coordinate, at every one of the $K-1$ non-trivial path points, at every pruning step, giving $\mathcal{O}(P)$ probes per step where the other schemes require $\mathcal{O}(1)$ evaluations.

\paragraph{Global path (\texttt{f\_dist\_global}).}
The expense of \eqref{eq:score_exact} is due entirely to the requirement that each coordinate be perturbed independently (a Fisher evaluation performed for coordinate $k$ tells us nothing about coordinate $k'$), which suggests a surrogate in which one family of models serves every coordinate at once.
Rather than shrinking one coordinate at a time, all are shrunk together, the metric being evaluated on the ray $\alpha\theta$ and every diagonal entry read off from each evaluation:
\begin{equation}
\label{eq:score_global}
s^{\rm global}_k
=
\big|\theta^k\big|\;\frac{1}{K}\sum_{\alpha}\sqrt{I_{kk}\big(\alpha\theta\big)}\,.
\end{equation}
The cost collapses from $\mathcal{O}(P)$ Fisher probes per pruning step to exactly $K$ Fisher evaluations, independent of the number of parameters (the difference, at the scales considered here, between days and seconds).

Two features of \eqref{eq:score_global} deserve comment.
First, the average is taken of the square root rather than of its argument, which is the literal discretisation of the length integral $\int\!\sqrt{g}\,dt$ and so is, in this respect, the better-motivated quadrature of the two; the exact measure \eqref{eq:score_exact} is nonetheless the better approximation overall, because it perturbs the correct path.
Second, its grid is not the one set out above but $\alpha \in \{1, \tfrac{K-1}{K}, \ldots, \tfrac{1}{K}\}$, excluding $\alpha = 0$: the all-zero model is degenerate, with $I_{kk}(0) \equiv 0$ identically, so including that point would contribute nothing but a rescaling of the one-shot score.
Since $\alpha \cdot 0 = 0$, already-pruned coordinates remain pruned at every $\alpha$, and the current sparsity mask is respected along the whole ray.

The approximation \eqref{eq:score_global} is uncontrolled in a way the others are not (it evaluates the metric on a ray through the origin rather than on the coordinate line) and there is no \emph{a priori} reason for it to track \eqref{eq:score_exact}.
Whether it does is an empirical question, and one of the questions \S\ref{sec:numerics} answers.

All of these approximation combinations are trialled in turn, alongside the magnitude and Fisher baselines of \S\ref{sec:prune}, to establish how closely each tracks the true Fisher distance in model space and what that fidelity is worth as a pruning criterion.

\section{Numerical Verification}\label{sec:numerics}

\subsection{Protocol}\label{sec:protocol}

The two architectures of \S\ref{sec:nns} are trained on each of the two datasets of \S\ref{sec:data}, giving four testbeds, and the whole experiment is repeated for five random seeds which control weight initialisation, data ordering, and the subsample on which the Fisher information is estimated.
Training details are common to all four combinations: $50$ epochs of \texttt{Adam} at learning rate $10^{-3}$ with batch size $64$, from scratch and without augmentation.
The resulting dense models are summarised in Table~\ref{tab:setup}; a separate checkpoint is trained for each seed, and every pruning scheme is applied to that same checkpoint, so that all comparisons within a seed are between prunings of one identical trained model.

\begin{table}[H]
\centering
\tabSetup
\caption{The four testbeds. Dense accuracy is on the test split, mean $\pm$ standard deviation over the five seeds. \texttt{SimpleNN}'s parameter count differs between datasets because the flattened input dimension does; \texttt{SimpleViT}'s differs only through the patch-embedding convolution.}
\label{tab:setup}
\end{table}

Each scheme is swept over the full range of pruning ratios $r \in \{0, 0.1, \ldots, 1.0\}$, where $r$ is the fraction of \emph{all} model parameters set to zero.
Pruning is unstructured and global, such that at each step the surviving parameters are ranked by the scheme's importance measure and the lowest-scoring are zeroed, with no retraining or fine-tuning at any point.
This last choice is deliberate.
Retraining after pruning recovers accuracy but confounds the quality of the importance measure with the capacity of the remaining network to relearn, and it is the measure alone that is under test here.
The sweep is run to $r = 1$ rather than stopping at a fixed sparsity budget, so that the comparison covers the full regime (including where the schemes substantially separate), rather than only the regime in which all of the schemes succeed.

Performance at each step is recorded as classification accuracy and as the Matthews correlation coefficient (MCC) \cite{Matthews:1975mcc}.
The latter is included because accuracy on a ten-class problem can be flattered by a degenerate predictor that has collapsed onto the majority class, whereas MCC is chance-corrected and reports such a collapse as zero \cite{Chicco:2020mcc}.
Both are normalised by their value on the unpruned model, so that curves from testbeds with different dense accuracies are commensurable, and each scheme is summarised by the area under its normalised curve over the sweep,
\begin{equation}
\label{eq:auc}
\mathrm{AUC} = \int_0^1 \! \mathcal{M}_{\rm norm}(r)\,dr\,,
\end{equation}
evaluated by the trapezoidal rule on the common grid.
A perfect scheme, losing nothing until the model is entirely removed, would score $1$.

A scheme that carried no information at all would not score $0$, however, but the value obtained by a model degraded to chance, which for normalised accuracy is $(1/C)/\mathcal{M}_{\rm dense}$, and so differs between testbeds of different dense accuracy.
Therefore the floor-corrected quantity $(\mathrm{AUC} - \mathrm{floor})/(1 - \mathrm{floor})$ is also reported, which rescales each testbed so that chance is $0$ and perfection is $1$, and is the appropriate figure for comparison \emph{across} architectures.
For MCC the floor is exactly $0$ (a collapsed one-class predictor has zero correlation with the labels) so the correction is the identity there and the column is omitted.

\subsection{Comparative Performance}\label{sec:results_main}

\begin{figure}[H]
\centering
\includegraphics[width=\textwidth]{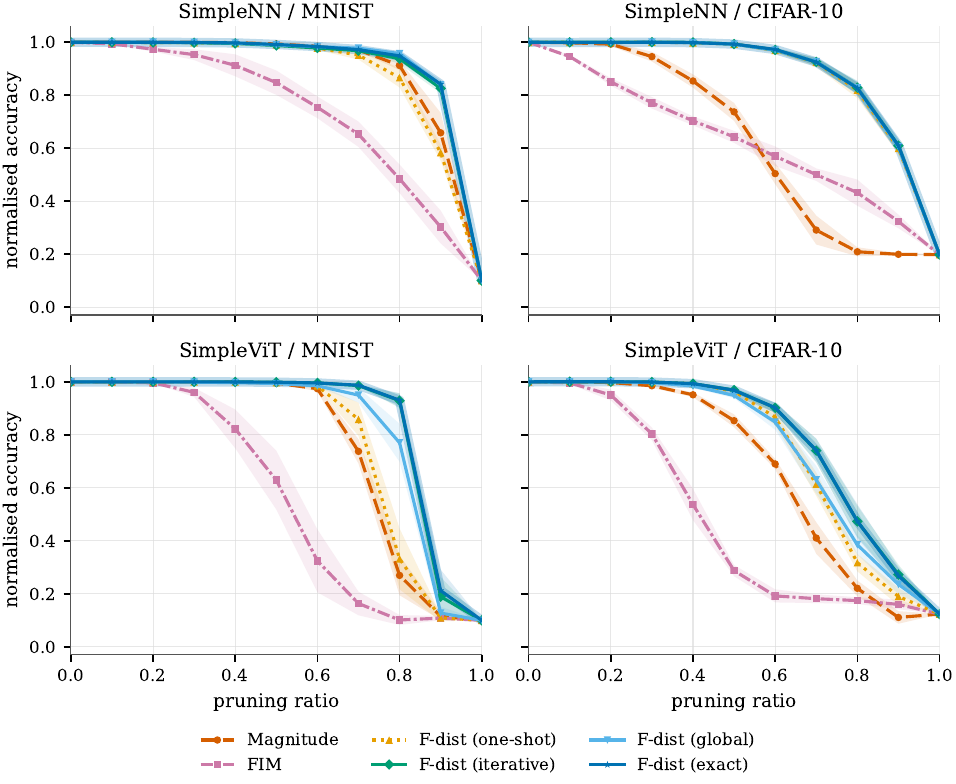}
\caption{Normalised test accuracy against pruning ratio for all six schemes on the four testbeds. Lines are means over five seeds and bands are $\pm$ one standard deviation. The exact Fisher-distance scheme is drawn beneath a wide translucent halo; where a cheaper scheme reproduces it, that curve is visibly contained within the halo (the iterative scheme is largely hidden beneath it on three of the four panels).}
\label{fig:curves}
\end{figure}

The principal results are collected in Table~\ref{tab:main}, and the underlying sweeps are shown in Figure~\ref{fig:curves}.

\begin{table}[H]
\centering
\footnotesize
\tabMain
\caption{Area under the normalised performance curve over the full $0$--$100\%$ pruning range, mean $\pm$ standard deviation over five seeds. The floor-corrected column rescales chance to $0$; it is omitted for MCC, whose floor is identically $0$. Compute is the median wall-clock of the pruning sweep over the five seeds. Here and in the figures, \emph{F-dist} abbreviates \emph{Fisher-distance}. $^\dagger$One \texttt{SimpleViT}/MNIST seed ran on a contended node and took $6.41$~d against $3.36$--$3.38$~d for the other four; the median is quoted rather than the mean, which that single outlier inflates by $18\%$.}
\label{tab:main}
\end{table}

These results show a clear superiority for the Fisher-distance family of pruning methods.
In each of the four testbeds, and on both metrics, the best-performing Fisher-distance scheme exceeds magnitude pruning: by $1.02\times$ on \texttt{SimpleNN}/MNIST, $1.41\times$ on \texttt{SimpleNN}/CIFAR-10, $1.14\times$ on \texttt{SimpleViT}/MNIST and $1.17\times$ on \texttt{SimpleViT}/CIFAR-10 in normalised accuracy, with the corresponding MCC ratios equal or larger throughout.
The advantage is largest where the task leaves room for it, where on \texttt{SimpleNN}/CIFAR-10 magnitude pruning has already lost half its accuracy by $r=0.6$ ($0.50$ of the dense value) while the Fisher-distance schemes are still at $0.97$ of it.

Importantly, from looking at the FIM scheme results, curvature alone is not the explanation.
The Fisher criterion \eqref{eq:fim_score} is \emph{worse} than magnitude pruning in three of the four testbeds ($0.82\times$, $0.74\times$ and $0.71\times$) and merely equal in the fourth.
On both transformer testbeds it is the weakest scheme tested by a wide margin. 
At $r=0.5$ it retains $0.63$ and $0.29$ of the dense accuracy on MNIST and CIFAR-10 respectively, where every Fisher-distance scheme is still above $0.94$.
This is worth emphasising, because it excludes the simplest reading of our results.
The Fisher-distance schemes do not succeed because second-order information is helpful; the standard second-order criterion actively harms performance here.
They succeed because the geometry is used to measure a \emph{displacement}, and displacement and curvature are only useful in combination.
The product structure of \eqref{eq:score_oneshot} is doing the work, not either factor alone.

Additionally, it is interesting to note that the MCC measure separates the schemes more sharply than accuracy.
On \texttt{SimpleNN}/CIFAR-10 the advantage over magnitude reads $1.41\times$ in accuracy but $1.58\times$ in MCC, and on \texttt{SimpleViT}/CIFAR-10 the Fisher criterion's deficit deepens from $0.71\times$ to $0.69\times$.
The ordering of schemes is essentially unchanged, which is the important check (the conclusions do not depend on the metric) but the chance-corrected measure consistently widens the gaps.
The interpretation is that the weaker schemes degrade in part by collapsing towards a majority-class predictor, a failure mode to which raw accuracy is partially blind and MCC is not.
The MCC sweeps are collected in Appendix~\ref{app:mcc}, where the correspondence is examined in more detail.

Finally, it is worth highlighting that the \texttt{SimpleNN}/MNIST testbed exhibits a ceiling effect.
There the advantage is only $1.02\times$, and \emph{one-shot} Fisher-distance pruning is actually worse than magnitude, at $0.98\times$.
Figure~\ref{fig:curves} shows why, magnitude pruning still retains $0.97$ of the dense accuracy at $r=0.7$ on this testbed, leaving almost no room in which a better measure could demonstrate itself, and the schemes separate only over the last two sweep points.
This is the ceiling anticipated in \S\ref{sec:data}, and it is the reason CIFAR-10 was included.
It is reported to illustrate that on a sufficiently easy problem the choice of importance measure does not matter much, and a method of this kind should be recommended for the regime where it does.

\subsection{The Approximation Hierarchy}\label{sec:results_hierarchy}

The hierarchy of \S\ref{sec:estimators} was constructed on the expectation that each refinement, being closer to the true model-space distance, would prune better.
That behaviour is not observed in full, and the manner in which it does not is highly informative for making sensible choices for efficient schemes.

\begin{figure}[H]
\centering
\includegraphics[width=\textwidth]{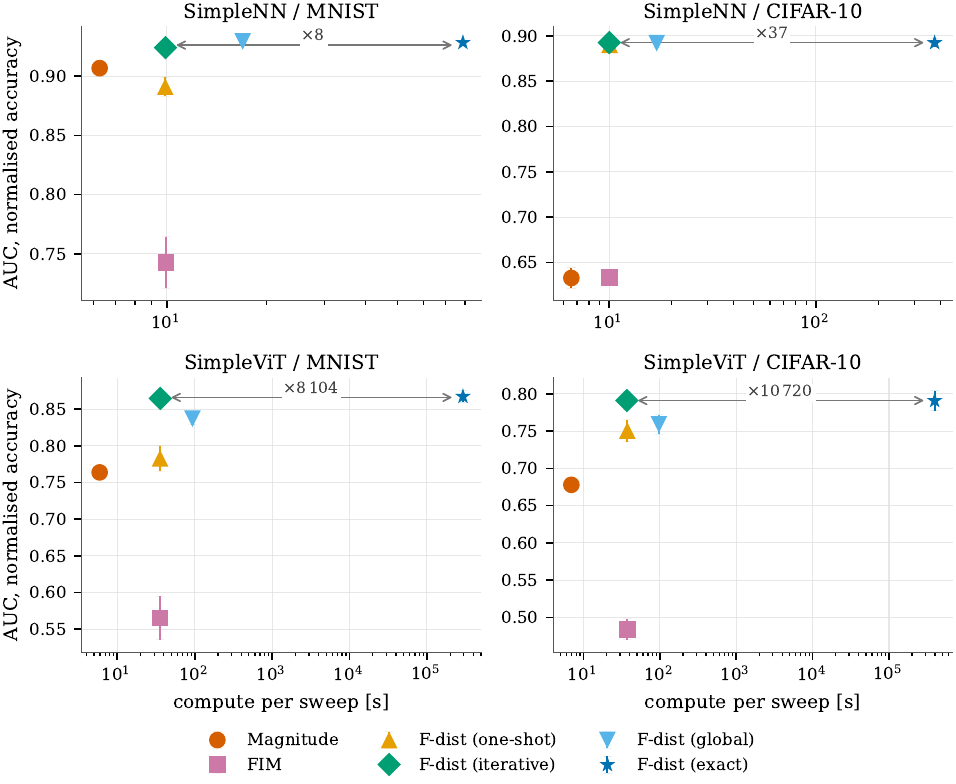}
\caption{Area under the normalised accuracy curve against the wall-clock cost of computing it, on logarithmic axes; error bars are $\pm$ one standard deviation over five seeds. The annotated arrow marks the cost ratio between the iterative and exact Fisher-distance measures, which on the transformer testbeds sit at the same height four decades apart.}
\label{fig:pareto}
\end{figure}

The one-shot measure \eqref{eq:score_oneshot} is indeed the weakest of the family in every testbed, and recomputing the metric at each step (the step from \eqref{eq:score_oneshot} to \eqref{eq:score_iterative}) is a genuine and sometimes large improvement, on \texttt{SimpleViT}/MNIST it moves the AUC from $0.783$ to $0.865$.
Resolving the metric along the path, however, buys essentially nothing beyond that.
Comparing the exact measure \eqref{eq:score_exact} with the iterative one \eqref{eq:score_iterative}:
\begin{center}
\small
\begin{tabular}{lcccc}
\toprule
& NN/MNIST & NN/CIFAR-10 & ViT/MNIST & ViT/CIFAR-10 \\
\midrule
$\mathrm{AUC}_{\rm exact} - \mathrm{AUC}_{\rm iter}$ & $+0.0043$ & $-0.0002$ & $+0.0026$ & $-0.0003$ \\
in units of $\sigma$ & $2.8$ & $0.05$ & $0.3$ & $0.02$ \\
cost ratio & $\times 8$ & $\times 37$ & $\times 8{,}104$ & $\times 10{,}720$ \\
\bottomrule
\end{tabular}
\end{center}
On three of the four testbeds the exact and iterative measures are statistically indistinguishable (differences of $0.02$ to $0.3$ standard deviations) while the exact measure costs between $37$ and $10{,}720$ times more to evaluate.
On the fourth, \texttt{SimpleNN}/MNIST, the exact measure is genuinely ahead of the iterative one by $2.8\sigma$; but on that same testbed the \emph{global} approximation \eqref{eq:score_global}, which is cheaper still, scores $0.930 \pm 0.002$ against the exact measure's $0.928 \pm 0.002$, and so is itself not beaten.
The correct statement is therefore the following: in every testbed, some member of the cheap family matches the exact measure.
Figure~\ref{fig:pareto} displays this directly.

The two cheap path-averaged schemes are not interchangeable, however, and the pattern of their difference is architectural.
On the fully-connected testbeds the global approximation \eqref{eq:score_global} is as good as anything else, and on \texttt{SimpleNN}/MNIST it is the best scheme overall.
On both transformer testbeds it falls clearly behind the iterative measure ($0.837$ against $0.865$, and $0.759$ against $0.791$) while also costing more.
This is attributed to the ray $\alpha\theta$ on which \eqref{eq:score_global} evaluates the metric: shrinking every parameter simultaneously is a mild perturbation of an \texttt{MLP}, but it drives a transformer's attention logits toward zero together, flattening the softmax and moving the model to a point whose Fisher information is not representative of any single-coordinate path.
The iterative measure \eqref{eq:score_iterative} makes no such excursion.

Taken together, these results recommend \eqref{eq:score_iterative} as the practical scheme.
It beats magnitude pruning in all four testbeds, matches or effectively matches the exact measure in all four, is the cheapest member of the Fisher-distance family\footnote{As noted in \S\ref{sec:estimators}, \eqref{eq:score_exact} averages inside the square root while \eqref{eq:score_global} averages outside it. Since the two agree to within seed noise on the testbeds where both are also close to the iterative measure, the variation of $I_{kk}$ along the path is evidently small enough that the distinction between the two quadratures is not resolved by these experiments.}, and is the only one whose behaviour does not depend on the architecture.
The value of the exact computation lies not in its use as a pruning criterion but in its role here as the ground truth which licenses the cheap one.

\subsection{Resolution of the Path}\label{sec:results_k}

The exact measure \eqref{eq:score_exact} carries one free parameter, the number $K$ of points at which the metric is resolved along the coordinate line.
Its cost is exactly proportional to $K-1$, the number of Fisher probes required per coordinate, so the question of how large $K$ must be is the question of what the scheme costs.

\begin{figure}[H]
\centering
\includegraphics[width=0.46\textwidth]{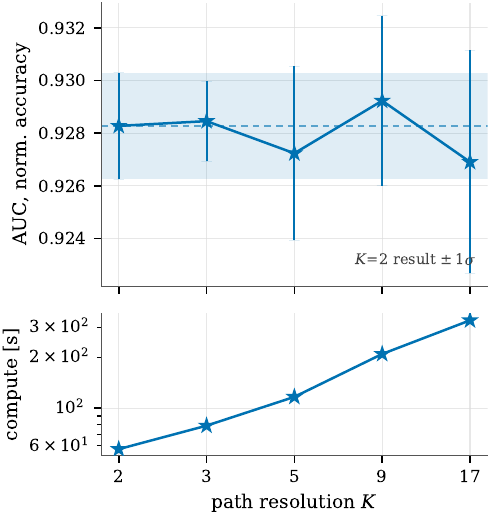}
\caption{Area under the normalised accuracy curve (above) and sweep cost (below) against path resolution $K$. The shaded band is the $K=2$ result $\pm$ one standard deviation; every richer grid lands inside it, while the cost rises by a factor of $5.8$.}
\label{fig:kladder}
\end{figure}

The ladder $K \in \{2, 3, 5, 9, 17\}$ is evaluated on \texttt{SimpleNN}/MNIST, the one testbed where the exact scheme is cheap enough to sweep repeatedly.
The ladder is chosen so that the grids are strictly nested ($\{1,0\} \subset \{1,\tfrac12,0\} \subset$ quarters $\subset$ eighths $\subset$ sixteenths) making it a genuine dyadic refinement, while the probe count per coordinate doubles exactly as $1,2,4,8,16$.
Results are given in Table~\ref{tab:k} and Figure~\ref{fig:kladder}.

\begin{table}[H]
\centering
\tabK
\caption{Dependence of the exact Fisher-distance measure on the path resolution $K$, for \texttt{SimpleNN}/MNIST over five seeds. The probe count per coordinate is $K-1$; compute is the median wall-clock of the full sweep.}
\label{tab:k}
\end{table}

The AUC is flat.
Across the whole ladder it spans $0.0023$, every point lies within $0.7$ standard deviations of the $K=2$ value, and the deviations alternate in sign (the signature of seed noise rather than of a trend).
Meanwhile the cost rises by a factor of $5.8$.
The crudest possible resolution of the path, $K=2$, using the single midpoint-free interval between the trained value and zero, is simultaneously the cheapest and the lowest-variance member of the ladder; sixteen-fold more probes per coordinate purchase nothing.

The implication is a statement about the geometry rather than about the algorithm: along the coordinate line from $(\theta^\ast)^k$ to zero, the Fisher information varies slowly enough that a two-point quadrature captures the integral to within the precision that pruning outcomes can resolve.
This is consistent with the finding of \S\ref{sec:results_hierarchy} that resolving the path at all is unnecessary, and it removes $K$ from the list of hyperparameters a practitioner must tune. 
It would be interesting to see whether this behaviour is still observed for larger architectures and on more complex tasks.

\subsection{Computational Cost}\label{sec:results_cost}

The compute column of Table~\ref{tab:main} spans six orders of magnitude and deserves comment, both because it is the quantity the preceding two subsections trade against accuracy and because the exact measure is only computable at transformer scale with some care.

The cheap schemes are all comparable. 
Magnitude pruning is free up to the cost of the sweep itself ($\sim 6$s), and the one-shot, iterative and Fisher schemes each add one Fisher evaluation per step, giving $10$s on \texttt{SimpleNN} and $36$s on \texttt{SimpleViT}.
The global scheme \eqref{eq:score_global} costs $K$ such evaluations, hence roughly threefold more.
The exact scheme \eqref{eq:score_exact} is in a different regime entirely, $\mathcal{O}(P)$ Fisher probes per pruning step, which at $P \sim 5\times10^5$ is some $10^7$ probes over a full sweep.

Two implementation results make this tractable.
First, the naive route (computing the full Fisher diagonal by back-propagation and reading off one entry) discards $P-1$ of the $P$ numbers it computes.
The identity
\begin{equation}
\label{eq:forward_fisher}
I_{ii}(\theta) = \mathbb{E}_{x}\,\mathrm{Var}_{c\sim p(\cdot\mid x,\theta)}\!\left[\frac{\partial f_\theta(x)_c}{\partial \theta^i}\right]
\end{equation}
delivers a single diagonal entry directly from one forward-mode derivative, at a measured speed-up of between $8.9\times$ and $325\times$ per probe depending on architecture and dataset.
This is an exact reformulation, not an approximation, and was verified against the back-propagation implementation to machine precision.
Second, the probes are mutually independent, so the coordinate list is sharded across processes; the sharding is arranged to preserve the serial batch shapes exactly, so that the result is bitwise identical to the serial computation rather than merely equal to within floating-point reordering.

With both in place, a full exact sweep for \texttt{SimpleViT} costs $3.4$--$4.6$ days on $128$ cores, and the complete set of exact transformer runs reported here amounts to approximately 40 CPU-node-days.
That figure is the price of the ground truth against which \S\ref{sec:results_hierarchy} calibrates the cheap schemes, and (given the conclusion reached there) it is a price that need be paid only once.

\section{Summary}\label{sec:summary}

Pruning a parameter moves a model a finite distance in its space of distributions, and the natural measure of that distance is supplied by information geometry.
Taking the Fisher information as the natural metric on model space and integrating along the path to $\theta^k = 0$ defines the importance of a parameter as its Fisher distance to zero.
The leading term of this quantity, obtained by assuming a diagonal metric, a coordinate-line path, and a constant metric along it, is the product $\sqrt{I_{kk}}\,|\theta^k|$ (exactly the two standard pruning criteria multiplied together).
Magnitude pruning and Fisher pruning are therefore not competing heuristics but complementary halves of one geometric object, each discarding what the other retains.

Relaxing the constancy assumption for the metric generates a hierarchy of measures, which were implemented in full (including the exact per-coordinate path integral) and evaluated across two architectures, two datasets, for the complete $0$--$100\%$ pruning range, and across five seeds.
The Fisher-distance family outperforms both baselines in all four testbeds and on both performance measures.
The Fisher criterion alone is worse than magnitude pruning in three of the four, which establishes that the improvement is due to the geometric combination rather than to inclusion of second-order information.
Furthermore, the refinements beyond recomputing the metric at each pruning step do not make a significant difference in these tasks.
In every testbed a cheap member of the family matches or exceeds the exact measure, at up to four orders of magnitude lower cost, while the resolution of the path may be reduced to its crudest two-point value with no measurable loss.
Our recommendation for practical use is accordingly the iterative measure \eqref{eq:score_iterative}, which is the cheapest, the most robust across architectures, and indistinguishable from the exact computation on every testbed examined in this work.

In terms of future developments, there are, excitingly, many potential avenues to further improve the geodesic distance approximation.
The most immediate concerns the diagonal approximation \eqref{eq:diag_approx}, which is the one assumption in the derivation that was not relaxed at any point.
The off-diagonal Fisher entries encode the correlations between parameters that unstructured pruning ignores when it removes weights independently, and a Kronecker-factored metric \cite{Martens:2015kfac} would make a block-structured Fisher distance computable at realistic scale.
It is a natural question whether the residual weakness on \texttt{SimpleNN}/MNIST reflects the ceiling of that testbed, as has been argued, or the limits of a diagonal metric.

A second direction concerns the path.
This work has assumed throughout that the infimum in \eqref{eq:distance_FIM_metric} is attained on the coordinate line, which is exact only if that line is a geodesic of the Fisher metric; in general it may not be.
The finding of \S\ref{sec:results_k} (that the metric varies too slowly along the path for its resolution to matter) suggests that the true geodesic departs little from the coordinate line in these models, but this is inference from a null result rather than a demonstration, and computing genuine Fisher geodesics for even a small network would provide informative evidence.

More broadly, the empirical fact that the crudest approximation performs comparably with the exact one is itself a question we should be able to theoretically answer.
Something about the Fisher geometry of trained networks makes the leading term unreasonably good, and identifying what that property is (whether its a property of the trained ensemble, of the loss landscape near a minimum, or of overparameterisation itself) would explain not only our results but the durable and somewhat puzzling competitiveness of the even cruder magnitude pruning \cite{Blalock:2020state}.
The connections between inference, renormalisation and information geometry developed in \cite{Berman:2022mak, Berman:2022uov, Berman:2023rqb, Berman:2024pax} offer one framework in which such a question might be posed, pruning being a coarse-graining of model space of exactly the kind that programme addresses.

Finally, to describe the practical applications and extensions.
Structured pruning, in which entire channels, heads, or blocks are removed, is what delivers wall-clock speed-ups on real hardware, and a Fisher distance to the submanifold on which a whole group of parameters vanishes is a well-posed object.
The schemes here prune without retraining in order to isolate the importance measure. One could instead combine the pruning with fine-tuning or with pruning-aware training and go futher than the current paradigm. 

Most importantly is that the compute result of \S\ref{sec:results_hierarchy} matters most where the benefits are the most valuable, i.e. at the scale of large pretrained models. A pruning criterion costing one Fisher evaluation per step is usable but one costing $\mathcal{O}(P)$ probes is less financially appealing. 

This work illustrates a methodology to achieve substantially more effective parameter pruning than the current state of the art. 
At scale, if a frontier LLM could be pruned with no (or negligible) performance loss then the inference costs could also be equivalently scaled down. 
The transformer models in this paper, on the more difficult datasets (ie. ViT on CIFAR-10), show a gap between magnitude pruning with no loss and F-dist pruning with no loss at  $\sim 10$\%.
This is an extraordinary financial statement. Assuming one can adapt architectures and hardware to exploit the weight pruning this would imply an approximate 10\% saving at inference time. 
Such a full 10\% saving is unlikely to be possible for frontier LLMs where the accepted performance loss will be very low and the architectural demands to exploit the weight pruning hard to achieve. However, the cash value of even a 1\% saving at inference would be estimated to be on the order of hundreds of millions of dollars a year.

Overall, this differential-geometric picture of a model's parameter space, through the naturally arising Fisher information metric, provides a means to compute meaningful measures of pruning effects on trained AI models. 
Given the challenge of AI sustainability, this research direction crucially provides an understanding of effective and theoretically justified methods for reducing AI inference costs.

\section*{Acknowledgements}
DSB acknowledges support from Pierre Andurand over the course of this research. 
EH is supported by S\~ao Paulo Research Foundation (FAPESP) grant 2024/18994-7; and thanks Alexander Stapleton and Marc Klinger for informative discussions.
This research utilised computational resources of the Centro Nacional de Processamento de Alto Desempenho
em S\~ao Paulo (CENAPAD-SP).

\section*{Data Availability}
The code and datasets used in this work are available here: \url{https://github.com/edhirst/Fdist_Pruning}. 

\newpage
\appendix
\section{Matthews Correlation Coefficient Results}\label{app:mcc}

\begin{figure}[H]
\centering
\includegraphics[width=\textwidth]{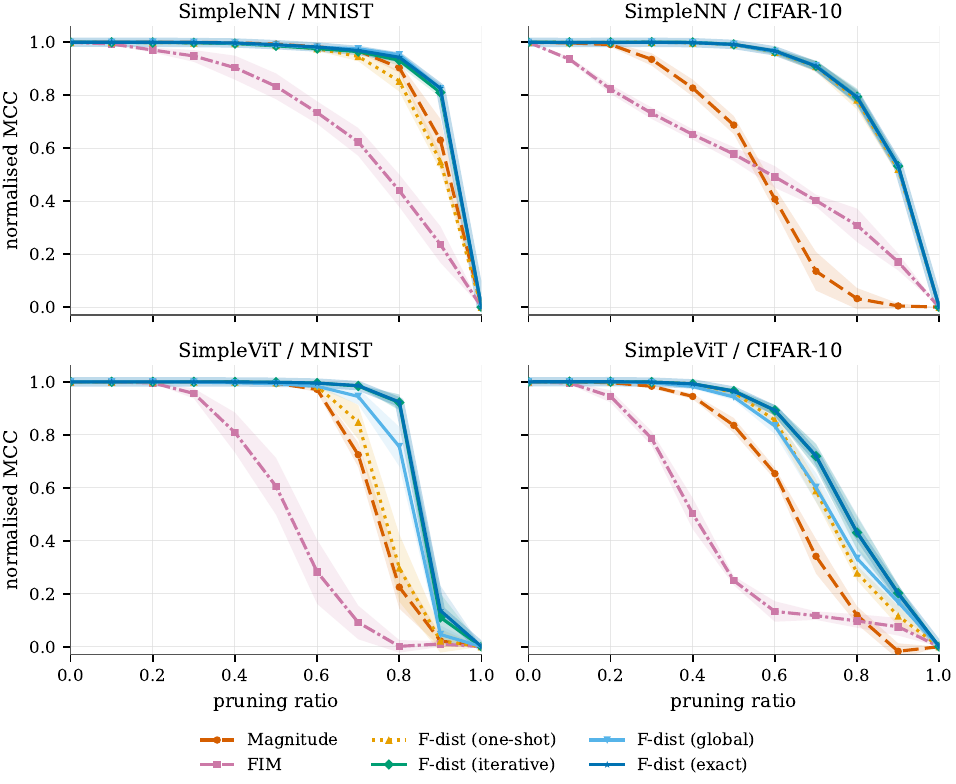}
\caption{Normalised Matthews correlation coefficient against pruning ratio, for all six schemes on the four testbeds. Lines are means over five seeds and bands are $\pm$ one standard deviation. Compare Figure~\ref{fig:curves}: the curves are qualitatively identical, displaced downwards by an amount which grows with the degree of degradation.}
\label{fig:curves_mcc}
\end{figure}

The Matthews correlation coefficient is reported alongside accuracy throughout \S\ref{sec:numerics} because a ten-class accuracy can be inflated by a predictor that has collapsed onto a single class, whereas MCC is chance-corrected and scores such a predictor at zero \cite{Matthews:1975mcc, Chicco:2020mcc}.
The AUC values are given in Table~\ref{tab:main}; the underlying sweeps are shown in Figure~\ref{fig:curves_mcc}, to be compared with Figure~\ref{fig:curves}.

The MCC results corroborate the conclusions of \S\ref{sec:results_main} in every respect, and sharpen two of them.

The ranking of schemes by AUC is identical under the two metrics in three of the four testbeds.
In the fourth, \texttt{SimpleViT}/CIFAR-10, the top two positions exchange (the exact measure edges ahead of the iterative one on MCC where the order is reversed on accuracy) but the two are statistically indistinguishable on both metrics there, scoring $0.770 \pm 0.015$ and $0.770 \pm 0.014$ respectively, so the exchange reflects seed noise rather than a genuine reordering.
No scheme changes rank against any scheme it actually separates from.
The choice of measure therefore does not affect any conclusion drawn in the main text.

The two sharpenings both concern the size of the gaps rather than their sign.
First, the advantage of the Fisher-distance family over magnitude pruning is larger on MCC in all four testbeds: $1.02\times \to 1.03\times$, $1.41\times \to 1.58\times$, $1.14\times \to 1.15\times$ and $1.17\times \to 1.21\times$.
Second, the deficit of the Fisher criterion deepens on the testbeds where it is already losing, from $0.82\times$ to $0.80\times$ on \texttt{SimpleNN}/MNIST, $0.74\times$ to $0.71\times$ on \texttt{SimpleViT}/MNIST and $0.71\times$ to $0.69\times$ on \texttt{SimpleViT}/CIFAR-10.

Both effects have the same explanation, and it is one that accuracy alone would have concealed.
A model degraded by a poor choice of pruning order does not fail by making uniformly random predictions; it fails by collapsing towards the majority class, retaining an accuracy near the class prior while carrying almost no information about the label.
MCC is sensitive to precisely this distinction and accuracy is not.
That the gaps widen under the chance-corrected metric means the schemes which lose are losing by more than their accuracy suggests, and confirms that the Fisher-distance criteria are preserving genuine discriminative structure rather than merely a favourable class balance.

\addcontentsline{toc}{section}{References}
\bibliographystyle{utphys}
\bibliography{references}{}

\end{document}